\documentclass[journal]{IEEEtai}

\usepackage[colorlinks,urlcolor=blue,linkcolor=blue,citecolor=blue]{hyperref}

\usepackage{color,array}

\usepackage{graphicx}

\usepackage{amsmath}
\usepackage{amssymb}
\usepackage{booktabs}
\usepackage{soul}
\soulregister\cite7
\soulregister\ref7
\soulregister\item7
\soulregister\itemize7
\soulregister\pageref7
\usepackage{multirow}
\usepackage{algorithm}
\usepackage{algorithmic}
\usepackage{caption}
\usepackage{subcaption}

\usepackage{amsfonts}
\usepackage{textcomp}
\usepackage{stfloats}
\usepackage{url}
\usepackage{verbatim}
\usepackage{graphicx}
\usepackage{cite}
\usepackage{tabularray}

\makeatletter 
\newcount\SOUL@minus
\makeatother  

\begin{document}

	\title{Spatiotemporal Object Detection for Improved Aerial Vehicle Detection in Traffic Monitoring}

	\author{Kristina Telegraph, Christos Kyrkou
		\thanks{Author Accepted Manuscript in Transactions on Artificial Intelligence. (\url{https://doi.org/10.1109/TAI.2024.3454566})}
		\thanks{The authors are with the KIOS Research and Innovation Center of Excellence, University of Cyprus.}}

	\maketitle
	
	\begin{abstract}
		This work presents advancements in multi-class vehicle detection using UAV cameras through the development of spatiotemporal object detection models. The study introduces a Spatio-Temporal Vehicle Detection Dataset (STVD) containing $6,600$ annotated sequential frame images captured by UAVs, enabling comprehensive training and evaluation of algorithms for holistic spatiotemporal perception. A YOLO-based object detection algorithm is enhanced to incorporate temporal dynamics, resulting in improved performance over single frame models. The integration of attention mechanisms into spatiotemporal models is shown to further enhance performance. Experimental validation demonstrates significant progress, with the best spatiotemporal model exhibiting a 16.22\% improvement over single frame models, while it is demonstrated that attention mechanisms hold the potential for additional performance gains.
	\end{abstract}
	
	\begin{IEEEImpStatement}
		Transportation systems have significant impacts on economic growth, social development and the environment. This article demonstrates how to better monitor road traffic using UAVs through a spatiotemporal deep learning model. This can provide the basis for the development of real-time traffic monitoring and control strategies, that when deployed can help to: 1) Perform real-time analysis of traffic patterns, congestion points, and bottlenecks at any location, and use this data to optimize signal timings, reroute traffic, and alleviate congestion. 2) Provide effective monitoring that helps detect incidents, such as accidents or stalled vehicles promptly, and help prevent secondary accidents. 3) Gain insights into usage pattern through long-term traffic monitoring data analysis, helping urban planners make informed decisions about road expansions, new infrastructure projects, and public transportation development.
	\end{IEEEImpStatement}
	
	
	\section{Introduction}
	\IEEEPARstart{V}{isual} object detection has taken enormous strides over the past decade thanks to advancements in deep learning architectures and availability of large datasets. These approaches have mainly focused on single image processing. Seminal works in this area include state-of-the-art image object detection methods such as the Region-based CNN series (RCNN) \cite{girshick_2014_rich, girshick_2015_fast, ren_2017_faster}, and the You Only Look Once series (YOLO) \cite{redmon_2016_you, redmon_2017_yolo9000, redmon_2018_yolov3, bochkovskiy_2020_yolov4, jocher_2022_ultralyticsyolov5,li2022yolov6, wang2023yolov7,yoloreview}, and more recently, transformer-based models such as the Detection Transformer (DETR) \cite{carion2020end,lv2023detrs} and the You Only Look at One Sequence (YOLOS) model \cite{fang2021you}, where by processing single images they are tasked with regressing bounding box information and associated classes. Hence, since videos are ultimately a sequence of image frames, they are also applicable to video processing by processing the incoming frame in isolation. The most popular techniques to incorporate temporal information and reasoning include post-processing methods such as tracking to build associations \cite{Pflugfelder2017SiameseVisualTracking}. As such, image-based detection on video-based data commonly encounters difficulties when dealing with phenomena such as occlusion, motion-blur, and variations in illumination conditions. Hence, it does not always produce reliable results, as each image will be perceived independently and it does not account for the aforementioned phenomena and valuable information present in the temporal domain. These drawbacks stem from the fact that the machine learning models cannot explicitly utilize the temporal information found in video streams. Hence, while generic image object detection has witnessed many achievements, there is still a gap in the utilization of both spatial and temporal information, which also requires appropriate datasets.
	
	\begin{figure}[!t]
		\centering
		\includegraphics[width=\columnwidth]{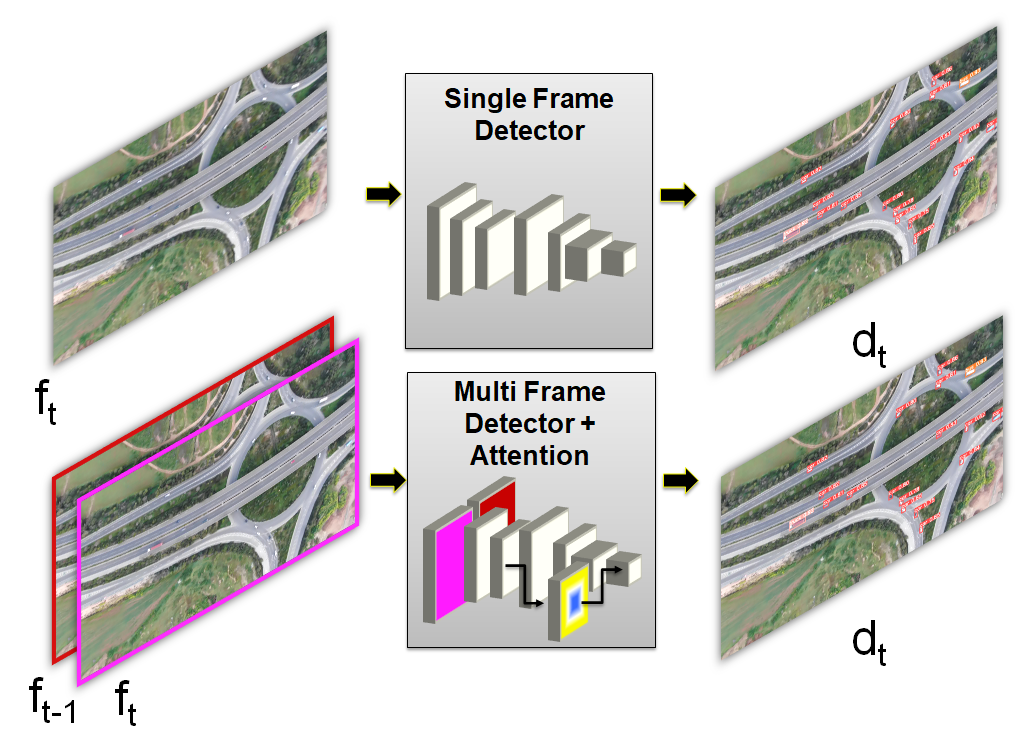}
		\caption{(top) Traditional Object Detection. (bottom) Proposed concept based on spatiotemporal models with attention.}
		\label{fig_concept}
	\end{figure}
	
	Promising results in generic image object detection led to the introduction of algorithms to extract motion information from videos. Initial approaches post-processed image detection algorithm results on video frames to enhance spatiotemporal cues \cite{han_2016_seqnms, kang_2017_object, kang_2018_tcnn}. In later works, researchers integrated these post-processing modules into the model itself to create end-to-end models for video object detection. These models employed different methods to capture the motion information, such as optical flow \cite{zhu_2017_flowguided, zhu_2017_deep}, and memory modules \cite{shi_2015_convolutional, lu_2017_online, ICDSC2018plastirasSTP}. Such techniques have not yet been investigated in the context of UAV-based sensing. UAVs are increasingly being used in transportation monitoring applications to provide on-demand and infrastructure-free information-rich (vehicle types, trajectories, counts, queue lengths) road traffic monitoring \cite{Makrigiorgis2022,Kyrkou2018,KyrkouPotentials2019}. Existing single frame models face challenges such as occlusions in dense regions, maintaining consistency of detection across frames, and have access to limited features of a single frame. 
	
	This work aims to enhance multi-class vehicle detection from UAV cameras through the development of spatiotemporal object detection models trained on 6-channel and 7-channel input data rather than conventional 3-channel RGB images (as shown in Fig. \ref{fig_concept}). By leveraging the temporal information through architectural modifications with different backbone and input options, we perform detection efficiently and in an end-to-end fashion that does not use optical flow or memory modules, and is suitable for real-time application. In particular, the main contributions of this article are summarized as follows:
	\begin{itemize}
		\item{A \textbf{S}patio-\textbf{T}emporal \textbf{V}ehicle \textbf{D}etection Dataset (STVD) comprising $6,600$ sequential frame images, meticulously annotated with categorizations encompassing three distinct vehicle classes: 'car', 'truck', and 'bus'. This dataset was created employing aerial footage acquired by Unmanned Aerial Vehicles (UAVs) across diverse segments of the road network situated within the geographical bounds of the Republic of Cyprus.} 
		\item{Investigate how to extend YOLOv5 object detection framework for processing of spatiotemporal data to encompass temporal dynamics through architectural enhancements and variations in input representations. Leading to improved performance over single frame models.}
		\item{We demonstrate that the introduction of attention mechanisms into the spatiotemporal models can lead to further performance improvements.}
	\end{itemize}
	
	Through a series of experiments we investigate different models, through both quantitative and qualitative analyses and examine class-specific performance. The results of the spatiotemporal models show significant progress with the best spatiotemporal model having 16.22\% improvement over the single frame model. Experiments also showed that incorporating attention-mechanisms into the spatiotemporal model architecture has potential to boost results even further.
	
	The remainder of the paper is organized as follows. Section II provides a summary of background information and related works in image and video object detection. Section III discusses the proposed approach and the custom dataset. Section IV describes the performance metrics considered and the setup of model training and inference experiments. Section V presents the experimental results, both quantitative and qualitative, while Section VI provides a discussion on the implications of the results. Finally, Section VII presents the main conclusions, as well as areas of improvement and future work.

	\section{Background and Relevant Work}
	\subsection{Image Object Detection}
	Image object detectors based on deep learning can be divided into two categories: two-stage detectors and one-stage detectors. An exemplary two-stage detector is the Faster R-CNN \cite{ren_2017_faster}, where candidate object bounding boxes are proposed in the first stage, and features are extracted from each candidate box in the second stage to carry out the bounding-box regression and classification tasks. These kinds of detectors possess very high accuracy at the expense of high inference speeds. One-stage detectors, on the other hand, such as SSD \cite{liu_2016_ssd} and the YOLO family of detectors\cite{redmon_2016_you, redmon_2017_yolo9000, redmon_2018_yolov3, bochkovskiy_2020_yolov4, jocher_2022_ultralyticsyolov5,li2022yolov6, wang2023yolov7,yoloreview}, do not have an initial region proposal step, they propose regressing bounding boxes and class probabilities from the images directly in one stage, making them more efficient and suitable for real-time applications due to their high inference speeds.
	
	YOLO \cite{redmon_2016_you} divides the input image into grid cells, where each grid cell is responsible for predicting the bounding box and confidence scores of objects centered within it. Confidence scores indicate how likely an object exists according to the model. YOLOv2 \cite{redmon_2017_yolo9000} further built on top of the original YOLO, adopting several novel concepts, such as anchor boxes and removal of fully connected layers to improve its speed and precision. The third generation of YOLO \cite{redmon_2018_yolov3} proposed a more robust feature extractor for its backbone called Darknet-53. It allowed adaptation to more complex datasets containing multiple overlapping labels. It also utilized three different feature map scales to predict bounding boxes, increasing its performance on smaller sized objects. YOLOv4 \cite{bochkovskiy_2020_yolov4} further added techniques to achieve the best speed-accuracy trade-off, through an improved loss function, Complete-IoU, and experimented with additional augmentation techniques. YOLOv5 \cite{jocher_2022_ultralyticsyolov5} focused on faster training and ease of use and demonstrated wide applicability across applications with a similar accuracy to YOLOv4.
	
	In more recent literature, transformer architectures have been explored in computer vision applications after achieving breakthroughs in areas of Natural Language Processing (NLP). The transformer architecture, which employs self-attention mechanisms, has proved to be effective in capturing long-range dependencies, prompting researchers to venture beyond just convolutions for image object detection and other vision tasks \cite{ramachandran2019stand}. Various studies have shown that employing both convolutions and self-attention in so-called hybrid CNN-transformer architectures can leverage both the local nature of convolutional layers and the global context that self-attention provides to achieve comparable and even better outcomes in practice \cite{bello2019attention},\cite{khan2023survey}. Overall, image object detection models lay the foundation for the needed development of video object detection networks \cite{jiao_2021_new}.
	
	\subsection{Video Object Detection}
	Video detection, or Spatiotemporal detection is a more challenging task, as it aims to detect patterns in both space and time. The earliest attempts to detect objects in video included using a state-of-the-art image detector on the video frames, extracting the spatiotemporal information, and using it to improve the image detector's preliminary results in a post-processing fashion. Sequence Non-Maximum Suppression (Seq-NMS) \cite{han_2016_seqnms}, Tubelet Proposal Networks (TPN) \cite{kang_2017_object} and Tubelets with CNNs (T-CNN) \cite{kang_2018_tcnn} all had a main strategy of mapping the results of image detectors across adjacent video frames, with the main difference between the post-processing methods being the mapping strategy used. While these methods are straightforward, post-processing techniques are usually undesirable and do not meet requirements for modern real-time applications. These methods are more time-consuming and usually not as effective due to their sequential nature. Hence, during inference, bounding boxes are processed and refined sequentially, rather than in one pass.
	
	Several methods for spatiotemporal understanding were later integrated into the single image detectors, allowing them to learn motion information directly during training in an end-to-end manner. Feature level methods that use optical flow such as Flow-Guided Feature Aggregation (FGFA) \cite{zhu_2017_flowguided} and Deep Feature Flow (DFF) \cite{zhu_2017_deep}, acquire temporal information from pixel-to-pixel correspondence between adjacent frames, using a key-frame to supplement features of other frames. However, adding optical flow to a network significantly increases model parameters, making these methods slow \cite{jiao_2021_new}.
	
	Subnetworks based on context were also integrated into single image detectors for the aim of spatiotemporal understanding. A variant of the traditional long short-term memory (LSTM) model \cite{hochreiter_1997_long}, that achieved very good results in many fields in the past, is the convolutional LSTM model \cite{shi_2015_convolutional}. It uses different ‘gate’ operations to extract and propagate features, thus it is able to establish context and long-term object associations between consecutive frames. The Association-LSTM \cite{lu_2017_online}, was proposed to improve video object detection. It consists mainly of the SSD \cite{liu_2016_ssd} image object detector, and a convolutional LSTM \cite{shi_2015_convolutional}. SSD performs the detection on each frame of the video, extracting individual frame features, which are then stacked and fed to the LSTM. Zhu and Liu \cite{liu2018mobile} introduced a lightweight model that also leveraged the combination of the SSD single image detector \cite{liu_2016_ssd} and convolutional-LSTM layers \cite{shi_2015_convolutional}. They inject the LSTM layers directly into a modified SSD detector to refine the input frame at each timestep and extract additional temporal cues. Zhang and Kim \cite{zhang2019modeling} proposed a temporal convolutional-LSTM approach that utilizes two types of temporal context information. Short-term context from adjusting the feature map from the directly preceding frame using optical flow, and long-term context from distant preceding frames through the convolutional LSTM. Comparisons with post-processing methods and flow-based methods showed improved results. Overall, while LSTMs require less computation than optical flow, they require more memory.

	Deng et al. \cite{deng2019object} attempted to tackle the storing of redundant information by exploiting external memory that comprises of addressable matrices, through attentional read and write processes. This allowed them to store information over a longer period, to be retrieved and aggregated through an attention-based global search. Beery et al. \cite{beery2020context} also utilized memory banks, both long-term and short-term, to store contextual information. Their proposed model, Context R-CNN, uses the two-stage Faster R-CNN architecture as the base model, wherein the first stage of detection, the box proposals of the network are routed through attention-based modules to incorporate temporal features from frames from the past. Another approach to extending convolutional neural networks into the time dimension is to employ 3-dimensional CNNs for spatiotemporal feature learning. Ji et al. \cite{ji_2013_3d} developed a 3D CNN model to capture motion information encoded in multiple adjacent frames for the task of action recognition in airport surveillance videos. Tran et al. \cite{tran_2015_learning} proposed 3D CNNs in the context of large-scale supervised learning tasks. They showed that 3D CNNs can outperform 2D CNNs on various video analysis applications. While 3D CNN based methods achieve good performance, their deployment is expensive as they have higher computational complexity than conventional 2D CNNs. Due to the robustness and efficiency of conventional CNNs, researchers were inclined to build onto the existing architecture merely by introducing a special module that can extract and learn temporal representations.  Lin et al. \cite{lin_2020_tsm} proposed the Temporal Shift Module (TSM), an approach that was able to achieve the performance of 3D CNNs without the added complexity and cost for action recognition. Passos et al. \cite{PASSOS2022101754} proposed a spatial-temporal consistency module that estimates the displacement of detected objects of interest from frame to frame. They extend the 2D spatial bounding boxes into the 3D space-time dimension by estimating an object's spatial displacement from one frame to another, aligning them space-wise and computing their pairwise intersection over union (IoU). 
	
	Spatiotemporal information was also utilized for the task of tiny object detection (TOD) in wide area motion imagery (WAMI). Lalonde et al. \cite{lalonde_2018_clusternet} proposed a two-stage spatiotemporal CNN that exploits both appearance and motion information using an input of five stacked grayscale frames utilizing a Faster-R-CNN-like region proposal network exceededing state-of-the-art results on the WPAFB 2009 dataset. Corsel et al. \cite{corsel_2023_exploiting} later outperformed this work on the same dataset using a spatiotemporal model based on the YOLOv5 object detection framework. They proposed two approaches, the first approach exploits temporal context by sampling every three consecutive frames from video sequences of the greyscale WPAFB 2009 WAMI dataset, where with each frame $f_t, f_{t-1}$ and $f_{t+1}$ were sampled with it, thus requiring a future frame to process the current frame. Their second approach was to use a two-stream architecture with the first stream handling the three frame representations from the first approach, and with the second stream handling exclusive motion information obtained from the absolute difference of the three frames used. They applied their models to single-class detection of tiny objects. Nevertheless, the necessity to wait for future frames to arrive makes this approach not suitable for real-time applications. Our work attempts to leverage temporal information on a smaller network, without sampling future frames, making it more suitable for real-time applications. Our work also applies the two-stream architecture approach using the absolute greyscale frame difference of two RGB frames, $f_t$, and $f_{t-1}$ along with the two frames themselves. We target multi-label, multi-class detection and classification on a custom aerial RGB dataset of road networks in complex settings captured from a UAV rather than from a single area in low resolution satelite images.
	
	\subsection{Attention Mechanisms}
	Attention in deep learning is a concept inspired by cognitive functions of humans, which is the natural tendency to selectively focus on parts of information deemed more important. It has been brought to the area of deep learning and computer vision with great success \cite{niu_2021_a}. The progress of attention-based models in computer vision in the deep learning era can be divided into four phases \cite{guo_2022_attention}. The first phase is characterized by work to combine deep neural networks with attention mechanisms, such as the RAM network \cite{mnih_2014_recurrent}, that recurrently predicts important features while updating the network concurrently. The second phase begins with the introduction of spatial attention, a mechanism that learns positions of interest in a spatial map. Jaderberg et al. \cite{jaderberg_2015_spatial} introduced a differentiable spatial transformer (STN) that finds these positions of interest through different transformations such as cropping, rotation, scaling and skew, adaptively according to the input feature map. The third phase began with a novel-attention mechanism called Squeeze-and-Excitation networks (SENet) \cite{hu_2019_squeezeandexcitation}. SENets adaptively recalibrated features channel-wise by explicitly modelling their interdependencies, to focus on the most important channels. More channel attention mechanisms followed, like Efficient Channel Attention networks (ECA-Net) \cite{wang_2020_ecanet} and Convolutional Block Attention module (CBAM) \cite{woo_2018_cbam}. The last and current phase of attention in computer vision is the self-attention era that was first introduced by Vaswani et al. \cite{vaswani_2017_attention} in transformers and rapidly revolutionized the field of natural language processing. The first to introduce self-attention to computer vision was Wang et al. \cite{wang_2018_nonlocal} who proposed non-local neural networks that showed superiority as they captured longer-range dependencies. The Detection Transformer (DETR) \cite{carion2020end} employed a CNN backbone to extract visual features and combined it with a Transformer-based decoder to perform object detection. It successfully combined the two in an end-to-end trainable pipeline. The Vision Transformer (ViT) \cite{dosovitskiy_2020_an} employed a multi-head self-attention architecture, and was able to achieve performance comparable to modern CNNs. A new branch of vision transformers called multi-scale vision transformers emerged where these models gradually reduced the number of tokens while increasing the number of token feature dimensions in a multi-stage hierarchical design, such as the Pyramid ViT \cite{wang2021pyramid}, and the Swin Transformer \cite{liu2021swin}. Multi-scale vision transformers were also extended and adapted for video understanding \cite{bertasius2021space,arnab_2021_vivit}, by perceiving videos as sequences of images that are similarly flattened into patches.
	
	\section{Proposed Approach}

	\subsection{\textbf{S}patio-\textbf{T}emporal \textbf{V}ehicle \textbf{D}etection Dataset (STVD)}
	\subsubsection{Dataset Contribution} In terms of datasets there has been considerable progression in static datasets \cite{cao2021visdrone}, multi-modal datasets \cite{DroneVehicle}, and altitude-aware datasets \cite{Makrigiorgis2023} for aerial object detection. However, these datasets are not suitable to investigate spatiotemporal detection models, since they do not have information from adjacent time instances. STVD is specifically designed to encapsulate both spatial and temporal information, as it consists of consecutive frames extracted from video clips. This temporal continuity offers a richer context for vehicle detection tasks, enabling algorithms to leverage motion dynamics and temporal dependencies.
	
	\begin{figure}[!t]
		\centering
		\includegraphics[width=3.4in]{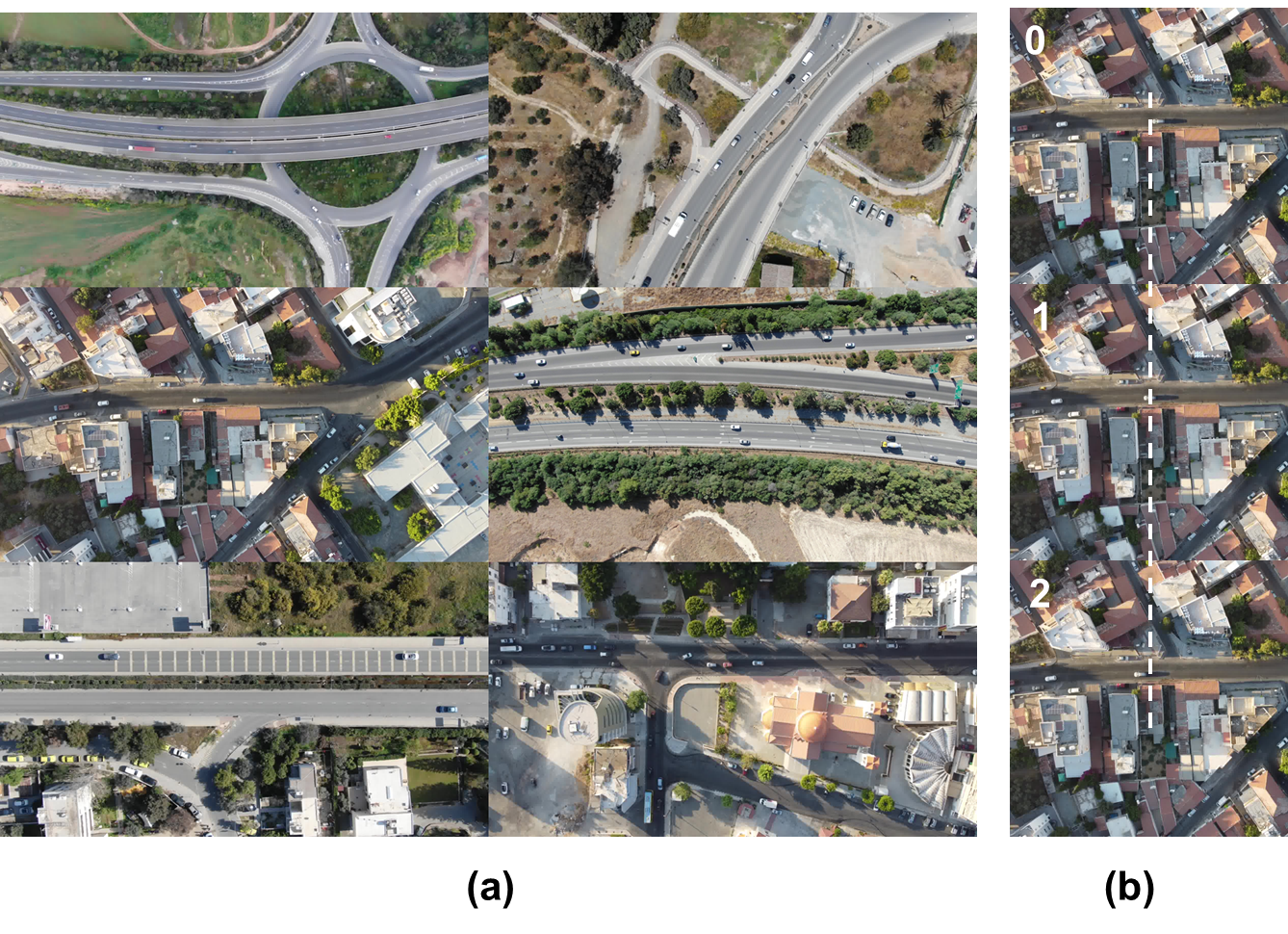}
		\caption{Illustrated images from the \textbf{S}patio-\textbf{T}emporal \textbf{V}ehicle \textbf{D}etection (STVD) Dataset. (a) The dataset covers different complex environments from different locations urban and rural. It has rounbabouts, highways and city environments with secondary roads. It also captures vehicles of varying sizes such as buses and normal passenger vehicles. (b) More importantly it captures the temporal flow for vehicle motion.}
		\label{fig_STVD}
	\end{figure}
	
	\begin{table}[t]
		\centering
		\caption{Dataset Details}
		\label{tab:dataset}
		\resizebox{\columnwidth}{!}{%
			\begin{tabular}{@{}ll@{}}
				\toprule
				\textbf{Dataset Feature} & \textbf{Description} \\ \midrule \midrule
				\textbf{Total Images} & $\sim$6600 \\
				\textbf{Image Sizes} & 1920x1080 \\
				\textbf{Classes} & \begin{tabular}[c]{@{}l@{}}3 classes\\ Car, Bus, Truck\end{tabular} \\
				\textbf{Data Collection} & Collect from UAVs at different locations in Nicosia, Cyprus\\
				\textbf{Data Format} & PNG \\
				\textbf{Labelling Format} & YOLO \\
				\textbf{Link} & \url{https://zenodo.org/records/11468690} \\ \bottomrule
			\end{tabular}%
		}
	\end{table}
	
	\subsubsection{Dataset Description}A dataset suitable for spatiotemporal object detection (summarized in Table \ref{tab:dataset} and visualized in Fig. \ref{fig_STVD}) is constructed using several aerial video clips of traffic in different road segments in Nicosia, Cyprus, captured using UAVs, rather than single areas in low resolution satelite images as other datasets. 
	A drone hovers statically over an area and captures the clips from a bird's eye view (top-down), directly on top of the street segment of interest, therefore there is no drone movement involved. The exact height of each video varies between $\sim80-110$ metres. Consequently, the object sizes also varied.
	
	By compiling multiple sequences of images extracted from these videos, the dataset accumulates a substantial corpus of $6,600$ frames. The dataset encapsulates 3 classes: ‘car’, ‘truck’ and ‘bus’ with a distribution of $81165$, $1541$, and $1625$ respectively in the case that we only use the even frame annotations, which approximately doubles when considering the entire dataset. An additional challenge of the dataset that mirrors real world application is the fact that the classes are not balanced, as there is a significantly larger number of cars compared to trucks and buses, as in a regular transportation network. The images have Full-HD resolution, with object sizes approximately between $20\times20$ to $150\times150$ pixels. The dataset was prepared in the YOLO format. The dataset was split into 80\% for training and the remaining 20\% for validation. The importance of such a dataset lies in its capability to encapsulate both spatial and temporal nuances. We note the frames belonging in the same continuous sequence as such the dataset can potentially be used to develop approaches that operate on multiple sequential frames for object detection by sampling a number of frames from the same sequence. While the dataset might seem small in comparison to other datasets every frame has a large number of vehicle instances which still enables learning rich representations. The naturally lower number of instances of 'truck' and 'bus' in a typical road network is also a challenge that our dataset replicates and is an important problem for the community to tackle.
	
	\subsubsection{Dataset Refinement} After the initial data collection phase, the collected images undergo a refinement to remove potential duplicate frames and bad-quality images. This refers to segments of clips where there was traffic at a standstill for example at a red traffic light, which results in a lot of frames that are almost the same, since cars do not move. In addition, frames depicting the same scenery over a few frames are removed to not bias the dataset. Footage that was not appropriate for our purpose i.e., during setup where the drone is moving and camera is not looking downwards was also removed. 
	
	\subsubsection{Data Annotation}The annotation process requires the rigorous labeling of images based on the different classes. Since a frame can contain hundreds of vehicles and annotation can be very time consuming we employ an accelerated auto-labelling process. First, we annotate part of the dataset and train an object detection model. This model is then used to provisionally predict bounding boxes of the remaining frames. Then all the frames were manually processed to account for misclassifications, missed detections, and double detections. This proved efficient, as they accurately placed bounding boxes on a significant portion of vehicle instances, requiring only minor adjustments. This enabled us to annotate a much larger set than previous works. To guarantee integrity of the results we employ a two-step approach, where a second independent researcher verified the annotated frames, ensuring a thorough validation procedure. It is worth noting that these single-image models were not employed further and did not influence any other processes or results. Finally, we take the first 80\% from every clip for the training set, and the rest for the validation set. This method of splitting mitigates this closeness in frames as the validation frames would be extracted from the end of the clips.
	
	\subsection{Detection Models}
	
	To develop spatiotemporal detection models we investigate how to better incorporate concepts and enhancements such as multi-frame processing and attention mechanisms to a baseline single frame detection network, namely YOLOv5 \cite{jocher_2022_ultralyticsyolov5}. This model also serves as a reference point for evaluating the effectiveness of the enhancements done to incorporate the temporal dimension. The YOLOv5 framework \cite{jocher_2022_ultralyticsyolov5}, was chosen for this work for its fast inference speeds, high level of accuracy and scalability. It is a member of the YOLO (You Only Look Once) family of single-stage regression object detectors, hence it uses a CNN architecture that is trained on single images and in one forward pass predicts object bounding boxes and their class probabilities. 
	
	\begin{figure}[!t]
		\centering
		\includegraphics[width=\columnwidth]{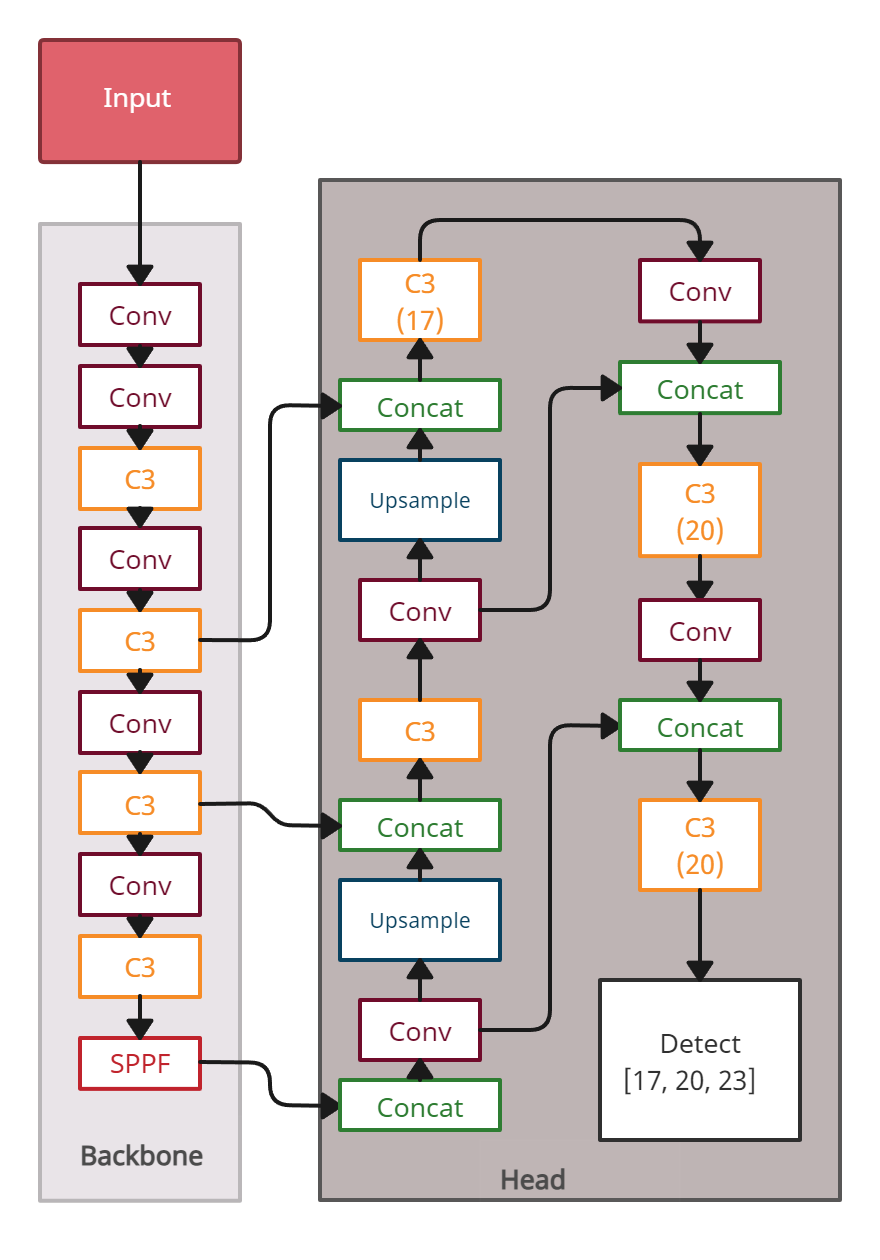}
		\caption{Single frame architecture based on YOLOv5.}
		\label{fig_1}
	\end{figure}
	
	The model architecture of the YOLOv5 model comprises of a backbone network, responsible for extracting deep-level features, and a head, which acts as a feature aggregator, combining features from the backbone from different scales, and a detection head which is responsible for making the predictions. The simplified architecture of YOLOv5 can be visualized in Fig. \ref{fig_1}. The architecture incorporated the C3 module, as seen in Fig. \ref{fig_1}, in both the backbone and head architecture, which is a simplified variant of the Cross Stage Partial network (CSP) block \cite{wang_2020_cspnet}.  YOLOv5 has multiple models of different sizes ranging from the smallest YOLOv5n, to the largest YOLOv5x, each designed for a specific use case of required speed-accuracy trade-off \cite{jocher_2022_ultralyticsyolov5}. In this work, the YOLOv5s architecture is adapted to create the models which provides a very good trade-off for real-time performance as well as accuracy and room for adding new features.
	
	\subsubsection{Single Frame Model}
	
\begin{figure}[!t]
		\centering
		\includegraphics[width=3.3in]{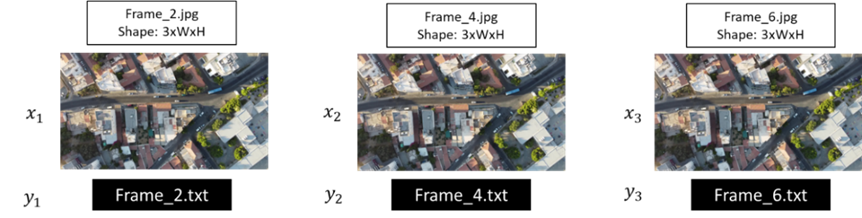}
		\caption{Overview of single frame model (even frames).}
		\label{fig_2}
	\end{figure}

	The single frame model is essentially the base YOLOv5s model that is trained on single frames to act as a benchmark for comparison with the spatiotemporal models and evaluate any improvements resulting from the added features. Fig. \ref{fig_2} provides an illustration of how this particular model processes examples independently of each other, as well as showing the input shapes corresponding to standard 3-channel images.
	
	\subsubsection{Spatiotemporal Models}
	A neural network exploiting both temporal and spatial relations of the input can be constructed in multiple ways. Most commonly, methods process spatial information first and fuse the output to add the temporal dimension. In addition, we also explore spatiotemporal models that employ additional temporal information which in this context is in the form of one previous frame that is processed by the same network from the beginning. The spatiotemporal models are trained on a sequence of paired images, utilizing the ground truth labels of the most recent frame for training and evaluation. Hence, since a pair acts as an evaluation instance, essentially, half the annotations are used. Thus for fairness, the same frames (half of the dataset) are used for training and evaluation of the single frame model as well. Next, we describe the different spatiotemporal models that are investigated.
	
\begin{figure}[!t]
		\centering
		\includegraphics[width=3.3in]{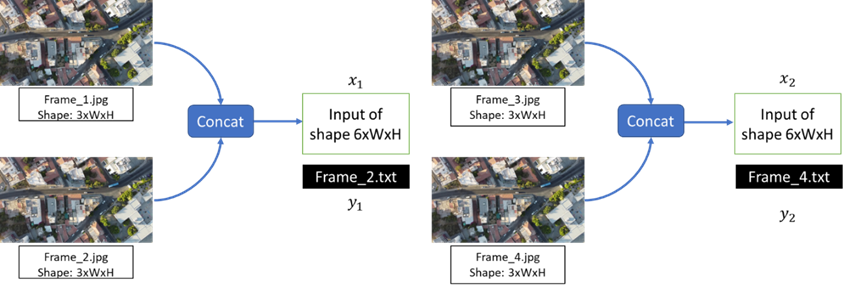}
		\caption{Overview of frame pair model.}
		\label{fig_3}
	\end{figure}

	\paragraph{Frame Pair Model}
	This model is a natural extension of a single frame model where the input is a pair of two consecutive frames at a time, concatenated channel-wise to result in a tensor of 6 channels.  Fig. \ref{fig_3} illustrates how the examples are combined and fed to the model. In this setting only the first convolutional layer is changed compared to the single frame to account for the larger channel size.
	
\begin{figure}[!t]
		\centering
		\includegraphics[width=3.3in]{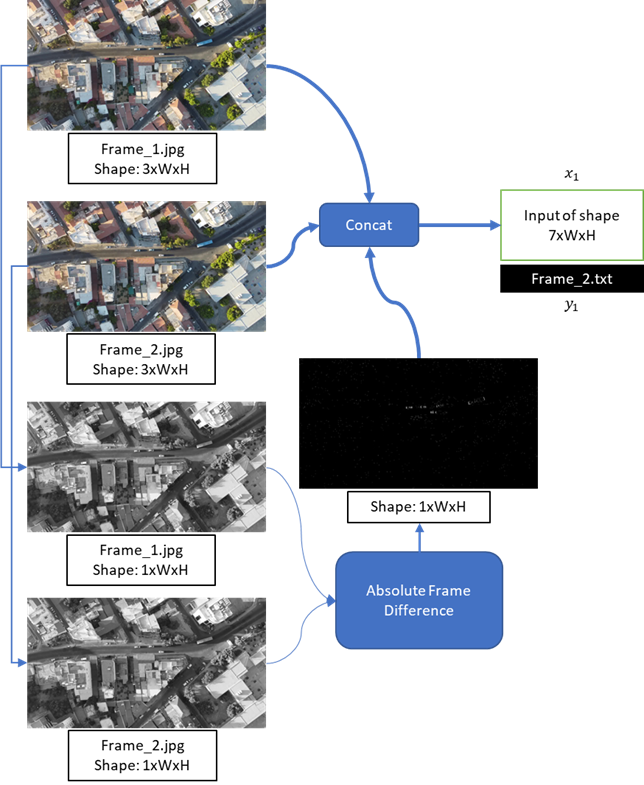}
		\caption{Overview of frame pair and difference model.}
		\label{fig_4}
	\end{figure}

	\paragraph{Frame Pair and Difference Model}
	This model naturally extends the pair model by adding an additional single-channel tensor which is the pixel-wise absolute greyscale frame difference of the two input images as shown in Fig. \ref{fig_4}, resulting in a 7-channel input. The main motivation behind this model being that we provide an additional signal that the model can use, that highlights areas of change and can thus guide the model to focus more on those regions that might either provide new information or be more challenging to detect (i.e., in the case of moving vehicles).
	
\begin{figure}[!t]
		\centering
		\includegraphics[width=\columnwidth]{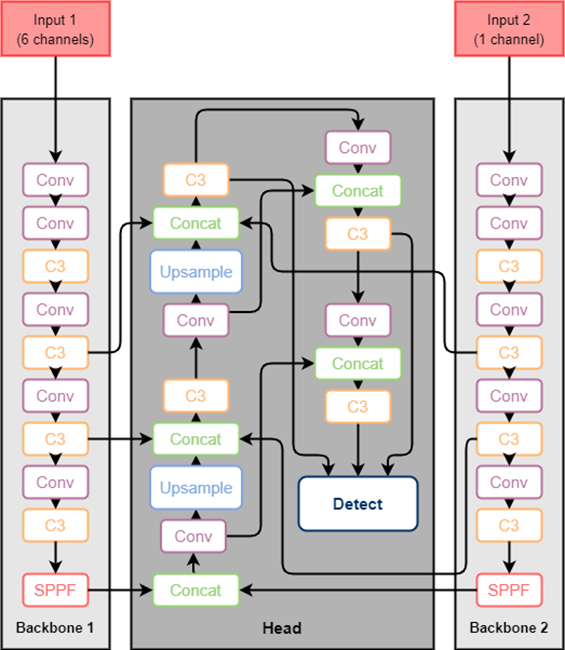}
		\caption{Two-stream model architecture.}
		\label{fig_5}
	\end{figure}

	\paragraph{Two-Stream Model}
	This model essentially uses the same input as the Frame Pair and Difference Model, however, the input is split in two as the frame pair and the frame difference channels are fed into two separate backbones, after which the outputs of both backbones are concatenated and passed to the model head, where the extracted features are concatenated at three detection scales. Both backbones are identical to that of the base YOLOv5 backbone shown in Fig. \ref{fig_1}, except for their input layer channels. The head architecture is also identical to that of base YOLOv5. Fig. \ref{fig_5} illustrates the proposed two-stream model architecture.
	
	\subsubsection{Attention-Spatiotemporal Models}
	The spatiotemporal models are further enhanced with attention mechanisms that provide the model with the flexibility of learning to which spatiotemporal information to attend to. The attention mechanisms would be applied to the spatiotemporal feature maps, allowing the model to refine them by focusing on the important information and ignoring the less important information. Two attention mechanisms were explored, Squeeze-and-Excitation networks (SENet) \cite{hu_2019_squeezeandexcitation}, as well as Efficient Channel Attention networks (ECA-Net) \cite{wang_2020_ecanet}. Their addition was investigated in both the Two-Stream Model as well as the Frame Pair and Difference Model.
	
	SENet \cite{hu_2019_squeezeandexcitation} is a CNN architecture that employs Squeeze-and-Excitation blocks. These blocks weigh their input channels adaptively according to their relevance, as opposed to convolutional layers in a CNN which give equal weights to each channel \cite{hu_2019_squeezeandexcitation}. ECA-Net \cite{wang_2020_ecanet} is a CNN architecture that employs Efficient Channel Attention blocks. As with SENets, ECA-Nets also provide channel attention but at a lower complexity trade-off and thus computational cost. It reduces each channel in the input tensor to a single pixel in the same way as in SENets, this vector is then subjected to a 1-D striding convolution. This makes ECA-Net more efficient as the total number of parameters added is just the size of the convolutional kernel, as opposed to SENets which employ a feed-forward network. By design, this also eliminates the dimensionality reduction present in SENets hidden layer. 
	
\begin{figure}[!t]
		\centering
		\includegraphics[width=\columnwidth]{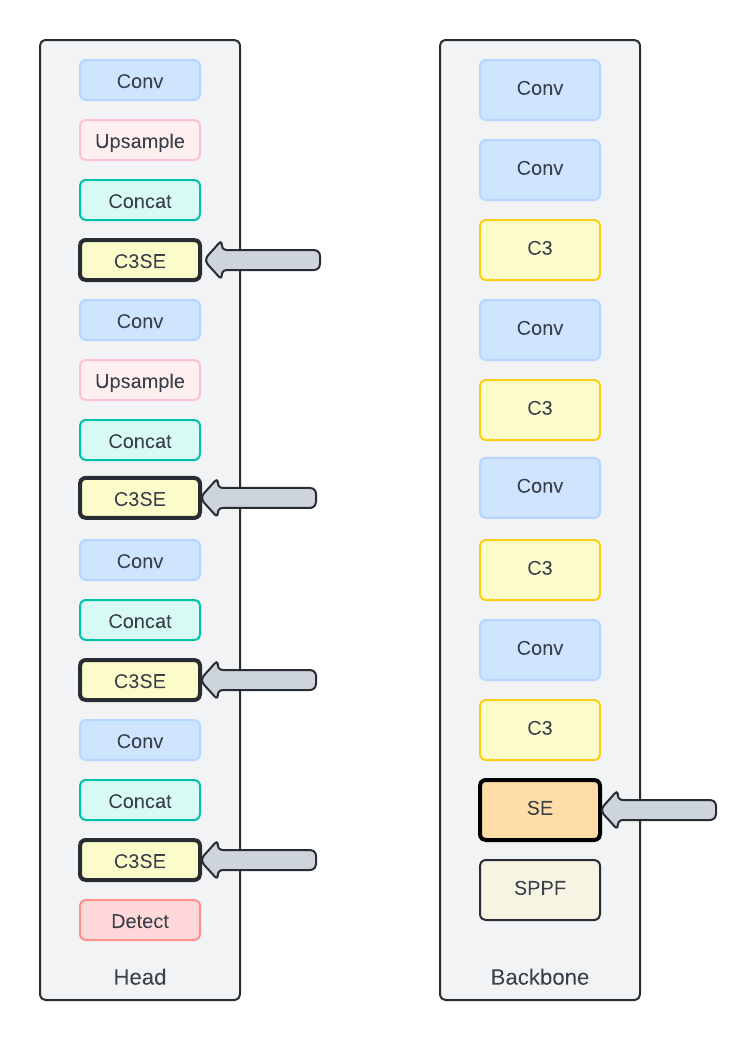}
		\caption{Two methods of adding attention mechanisms: embedded as a layer in spatiotemporal models' backbone (right), attention embedded within C3 modules in spatiotemporal models’ head (left).}
		\label{fig_6}
	\end{figure}

	The addition of these blocks to the spatiotemporal models was explored in two different ways in the model architecture, adding them as a layer at a single level at the end of the backbone, right before the final spatial pyramid pooling layer, as well as embedding them into the C3 modules at four different levels in the head architecture. In the attention embedded two-stream model, the attention layers are added in both backbones 1 and 2, at the same position. As a result, the two-stream-SE model has two SE layers, one in each backbone. Similarly, the two-stream-ECA model has two ECA layers, one in each backbone. Another approach that was explored was embedding the SE and ECA mechanisms into the C3 modules in the head architecture to apply attention during feature aggregation at different scales. The C3 module discussed earlier, is responsible for extracting feature maps at different scales and resolutions. Therefore, adding attention mechanisms in the C3 blocks could help the network amplify the more important features at different input scales and resolutions. The resulting modules of embedding SE and ECA into the C3 modules were called the C3SE and C3ECA modules respectively. Fig. \ref{fig_6} illustrates the two methods of adding attention mechanisms to the architecture. The backbone column (right) shows where the SE or ECA layers are added in the first method, and the head column (left) shows how these modified C3SE modules are substituted in place of the original C3 modules in the head architecture of the spatiotemporal models. In the attention-spatiotemporal models that utilize C3ECA modules, they are also implemented in the same way.
	
	\section{Training and Experiments Setup}
	
	\subsection{Setup}
	\subsubsection{Training}
	Training spatiotemporal models can pose several challenges compared to standard single image models since we no longer can retrieve single images completely at random. Thus, for training the model we implement appropriate data loading mechanisms that retrieve only valid pairs of frames for training to respect the temporal frame ordering. In addition, frames corresponding to different video locations are also not mixed together. Compared to default training strategies used in YOLO models the following adjustments were made. 1) Mosaic augmentation is disabled since they result in drastic reduction in the size of the objects which are already small, but more importantly, they would also mix frames from different times and locations. 2) For the spatiotemporal models any augmentation that randomizes the frame order was also disabled since the ordering needs to be preserved so that the temporal domain information remains valid. Instead, images are shuffled during training in the single frame model, and are shuffled in pairs before training in the spatiotemporal models. Traing hyperparameters are shown in Table \ref{tab:table6}.

	\begin{table}[]
		\centering
		\caption{Algorithm \& Training hyperparameters}
		\label{tab:table6}
		\begin{tabular}{@{}ll@{}}
			\toprule
			Parameter & Value \\ \midrule \midrule
			learning rate & 0.01 \\
			momentum & 0.937 \\
			weight decay & 0.0005 \\
			box loss gain & 0.05 \\
			cls loss gain & 0.5 \\
			obj loss gain & 1 \\
			IoU threshold & 0.6 \\
			anchor multiple threshold & 4 \\ \bottomrule
		\end{tabular}
	\end{table}
	
	\subsubsection{Validation and Inference}
	We derive the best model from training by picking the one that performed best on a weighted combination of the 10\% mAP@0.5 and 90\% mAP@0.5:0.95, on the validation set and then used it for inference testing.
	The image size used was $640$ as with training, and the batch size was $32$. The results also contained the class-specific metrics. Inference tests were carried out to perform a qualitative analysis of the spatiotemporal as well as the attention-spatiotemporal models’ results. The confidence threshold for the inference tests was $0.25$. The tests were done on several video frames from the validation set of different road segments and environments to also measure inference speed on a NVIDIA Tesla V100 GPU.
	
	\subsection{Performance Metrics}
	
	We monitor different metrics, summarized below to provide a comprehensive evaluation of the different models and the effect of each addition.
	
	\subsubsection{Precision and Recall}
	Precision is the percentage of correct positive predictions. It is the model’s ability to identify and detect only relevant objects. Recall is the model’s ability to find all relevant objects, it is the percentage of correct positive predictions among all ground-truths \cite{padilla_2021_a}.
	In order to establish prediction ‘correctness’, the measurement of intersection over union (IOU) is used, which measures the area overlap between the predicted bounding box and the ground-truth bounding box divided by the union area between them \cite{padilla_2020_a}.
	
	To calculate precision and recall, the following concepts also need to be defined:
	\begin{itemize}
		\item{True positive (TP): Correct detection of a ground-truth bounding box that exceeds an IoU threshold.}
		\item{False positive (FP): Incorrect detection of an object that is nonexistent, or a misplaced detection of an already matched ground truth object.}
		\item{False negative (FN): Undetected ground-truth box.}
	\end{itemize}
	
	Given each detected bounding box is classified as one of the above, precision (P) and recall (R) can be formally expressed by $P = TP/(TP + FP)$ and $R = TP/(TP + FN)$ respectively \cite{padilla_2020_a}.

	\subsubsection{Mean Average Precision}
	Average precision is a metric based on the area under the precision/recall curve, however, as this curve often has a zig-zag nature, it is first processed to smooth it out. Mean average precision is the mean over all classes of the dataset \cite{padilla_2020_a}. The \textit{mAP50} metric is the main performance metric used to compare the models, which entails that a predicted box is considered correct if the IoU with the ground truth box is greater than or equal to $0.5$.
	
	\subsubsection{Inference Speed}
	The model inference speed is a metric to consider if the model is aimed to be used in real-time applications. The speed is recorded for all models during inference experiments, indicated as the inference time per frame, as well as in terms of frames per second (FPS). Generally, the higher FPS the better as the model is then capable to process frames from multiple input sources (i.e., video streams) if needed.
	
	\section{Results}
	\subsection{Static Model Investigation}
	To investigate if the choice of model would have any impact on the performance, we trained both YOLOv5, YOLOv8 as well as RT-DETR \cite{lv2023detrs} models on a static (i.e., no temporal information) aerial dataset \cite{Makrigiorgis2022} for single frame processing. Results are shown in Table \ref{tab:YOLOcomp}.
	
	\begin{table}[h]
		\centering
		\caption{Comparison of YOLOv5, YOLOv8, RT-DETR}
		\label{tab:YOLOcomp}
		\resizebox{\columnwidth}{!}{
			\begin{tabular}{lcccc}
				\hline
				Model & $mAP\_0.5$ & $mAP\_0.5:0.95$ & Inference Speed (ms) & GFLOPS \\
				YOLOv5 & 0.946 & 0.682 & 4.3  & 47.9 \\
				YOLOv8 & 0.937 & 0.678 & 12.3 & 78.7 \\
				RT-DETR & 0.894 & 0.566 & 85.7 & 103.4 \\
				\hline
			\end{tabular}
		}
	\end{table}
	
	The results demonstrate that the performance of the two YOLO medium-size variants is very close in terms of detection accuracy, for both MAP scores. While YOLOv5 is overall more efficient which is important for real-time resource constrained applications. As a result, it is not anticipated that there will be a major difference when choosing alternative backbones. Furthermore, as the approaches introduced in this work are agnostic to the underlying convolutional architecture and training regime they can be adapted without much effort to any detection model, but this goes beyond the scope of this work. This is in line with recent studies in \cite{yoloreview} where it is observed that YOLOv5 provides very good accuracy-computation trade-offs against more recent versions, which makes it very suitable for our purposes. With regards to transformer-based models such as RT-DETR our results indicate that compared with existing YOLO models, they still lag behind when it comes to accuracy of small objects, a limitation also shown in \cite{lv2023detrs}, while also being slower at inference time and requiring more FLOPS.
	
	\subsection{Spatiotemporal Model Validation}
	\subsubsection{Spatiotemporal Models}
	Table \ref{tab:table2} shows the validation results of the single frame and the three spatiotemporal models for each and all classes of the dataset.
	
	\begin{table}[]
		\centering
		\caption{Validation results of single frame and spatiotemporal models}
		\label{tab:table2}
		\resizebox{\columnwidth}{!}{%
			\begin{tabular}{@{}ccllll@{}}
				\toprule
				\multicolumn{1}{l}{Input} & \multicolumn{1}{l}{Model} & Class & P & R & mAP50 \\ \midrule \midrule
				\multirow{4}{*}{Single Frame} & \multirow{4}{*}{Original} & All & 0.777 & 0.697 & \textbf{0.715} \\
				&  & Car & 0.977 & 0.940 & 0.969 \\
				&  & Truck & 0.816 & 0.375 & 0.484 \\
				&  & Bus & 0.538 & 0.777 & 0.691 \\ \midrule
				\multirow{12}{*}{\begin{tabular}[c]{@{}c@{}}Multi-Frame\\ (Spatiotemporal)\end{tabular}} & \multirow{4}{*}{Frame Pair} & All & 0.856 & 0.736 & \textbf{0.778} \\
				&  & Car & 0.986 & 0.929 & 0.972 \\
				&  & Truck & 0.89 & 0.448 & 0.532 \\
				&  & Bus & 0.691 & 0.832 & 0.832 \\ \cmidrule(l){2-6}
				& \multirow{4}{*}{Frame Pair \& Difference} & All & 0.79 & 0.752 & \textbf{0.811} \\
				&  & Car & 0.979 & 0.947 & 0.976 \\
				&  & Truck & 0.926 & 0.494 & 0.734 \\
				&  & Bus & 0.465 & 0.817 & 0.722 \\ \cmidrule(l){2-6} 
				& \multirow{4}{*}{Two Stream} & All & 0.906 & 0.781 & \textbf{0.831} \\
				&  & Car & 0.979 & 0.944 & 0.972 \\
				&  & Truck & 0.881 & 0.558 & 0.626 \\
				&  & Bus & 0.859 & 0.842 & 0.895 \\ \bottomrule
			\end{tabular}%
		}
	\end{table}
	
	It can be first observed from Table \ref{tab:table2} that all three spatiotemporal models result in improved mAP50 over the single frame model, with the two-stream model having the highest overall performance improvement. The frame pair model achieves a total mAP of 0.778 on all three classes, an 8.81\% improvement over the single frame model. The frame pair and difference model achieves an improvement of 13.43\% with a total mAP of 0.811. Finally, the two-stream model achieves the highest total mAP of all classes, with a total improvement of 16.22\% over the single frame model.
	Comparing the performance of the models in different classes, all spatiotemporal models show slightly improved mAP50 on the ‘car’ class. For the ‘car’ as well as the ‘truck’ class, the frame pair and difference model gives the highest mAPs of 0.976 and 0.734 respectively. The frame pair model and the two-stream model present a smaller improvement in the ‘truck’ class over the single frame model’s very low mAP of 0.484. The ‘bus’ class mAP performance is also improved in all three spatiotemporal models, with the Two-Stream model indicating the largest improvement over the single frame model from 0.691 to 0.895 mAP.
	To sum up, the spatiotemporal models exhibit improvement in results for all classes, and significant improvement for the minority classes of ‘truck’ and ‘bus’ that exhibited very poor performance in the single frame model.
	
	\subsubsection{Attention-Spatiotemporal Models}
	Tables \ref{tab:table3} and \ref{tab:table5} illustrate the validation results of the attention infused two-stream and frame pair and difference spatiotemporal models for each and all classes of the dataset, alongside the results of the regular relevant spatiotemporal models to form a comparison.
	
	\begin{table}[]
		\centering
		\caption{Validation results of attention two-stream spatiotemporal models}
		\label{tab:table3}
		\resizebox{\columnwidth}{!}{%
			\begin{tabular}{@{}ccllll@{}}
				\toprule
				\multicolumn{1}{l}{Input} & \multicolumn{1}{l}{Model} & Class & P & R & mAP50 \\ \midrule \midrule
				\multirow{4}{*}{Plain Spatiotemporal} & \multirow{4}{*}{Two Stream} & All & 0.906 & 0.781 & \textbf{0.831} \\
				&  & Car & 0.979 & 0.944 & 0.972 \\
				&  & Truck & 0.881 & 0.558 & 0.626 \\
				&  & Bus & 0.859 & 0.842 & 0.895 \\ \midrule
				\multirow{16}{*}{Spatiotemporal + Attention} & \multirow{4}{*}{Two Stream - SE} & All & 0.853 & 0.779 & \textbf{0.844} \\
				&  & Car & 0.985 & 0.924 & 0.973 \\
				&  & Truck & 0.887 & 0.561 & 0.712 \\
				&  & Bus & 0.688 & 0.851 & 0.848 \\ \cmidrule(l){2-6} 
				& \multirow{4}{*}{Two Steam - ECA} & All & 0.952 & 0.788 & \textbf{0.833} \\
				&  & Car & 0.985 & 0.938 & 0.975 \\
				&  & Truck & 0.954 & 0.551 & 0.616 \\
				&  & Bus & 0.917 & 0.876 & 0.910 \\ \cmidrule(l){2-6} 
				& \multirow{4}{*}{Two Stream - C3SE} & All & 0.864 & 0.771 & \textbf{0.861} \\
				&  & Car & 0.984 & 0.936 & 0.972 \\
				&  & Truck & 0.942 & 0.554 & 0.771 \\
				&  & Bus & 0.667 & 0.822 & 0.839 \\ \cmidrule(l){2-6} 
				& \multicolumn{1}{l}{\multirow{4}{*}{Two Stream - C3ECA}} & All & 0.874 & 0.766 & \textbf{0.839} \\
				& \multicolumn{1}{l}{} & Car & 0.984 & 0.935 & 0.973 \\
				& \multicolumn{1}{l}{} & Truck & 0.908 & 0.505 & 0.652 \\
				& \multicolumn{1}{l}{} & Bus & 0.731 & 0.856 & 0.892 \\ \bottomrule
			\end{tabular}%
		}
	\end{table}
	
	It can be seen from Table \ref{tab:table3} that all attention models show higher performance at varying degrees over the standard two-stream model. With the two-stream–C3SE–head model having the highest overall increase in performance of 3.61\% over the standard model, due to the steep increase in the mAP50 on the ‘truck’ class, regardless of the slight decrease in performance on the ‘bus’ class.
	
	\begin{table}[]
		\centering
		\caption{Validation results of attention frame pair and
			difference spatiotemporal models}
		\label{tab:table5}
		\resizebox{\columnwidth}{!}{%
			\begin{tabular}{@{}ccllll@{}}
				\toprule
				\multicolumn{1}{l}{Input} & \multicolumn{1}{l}{Model} & Class & P & R & mAP50 \\ \midrule \midrule
				\multirow{4}{*}{Plain Spatiotemporal} & \multirow{4}{*}{Frame Pair \& Difference} & All & 0.790 & 0.752 & \textbf{0.811} \\
				&  & Car & 0.979 & 0.947 & 0.976 \\
				&  & Truck & 0.926 & 0.494 & 0.734 \\
				&  & Bus & 0.465 & 0.817 & 0.722 \\ \midrule
				\multirow{8}{*}{Spatiotemporal + Attention} & \multirow{4}{*}{Frame Pair \& Difference - SE} & All & 0.818 & 0.764 & \textbf{0.844} \\
				&  & Car & 0.981 & 0.934 & 0.972 \\
				&  & Truck & 0.982 & 0.432 & 0.679 \\
				&  & Bus & 0.490 & 0.926 & 0.880 \\ \cmidrule(l){2-6} 
				& \multirow{4}{*}{Frame Pair \& Difference - C3SE (Head)} & All & 0.808 & 0.770 & \textbf{0.819} \\
				&  & Car & 0.970 & 0.955 & 0.976 \\
				&  & Truck & 0.869 & 0.519 & 0.640 \\
				&  & Bus & 0.585 & 0.837 & 0.841 \\ \hline
			\end{tabular}%
		}
	\end{table}

	\begin{figure*}[!t]
		\centering
		\subfloat[]{\includegraphics[width=2.5in]{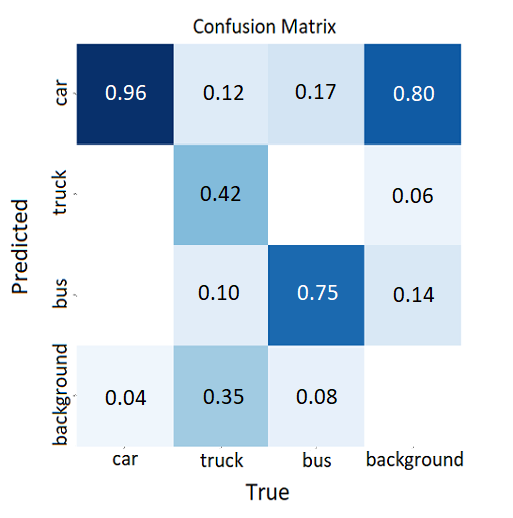}%
			\label{fig8_a}}
		\hfil
		\subfloat[]{\includegraphics[width=2.5in]{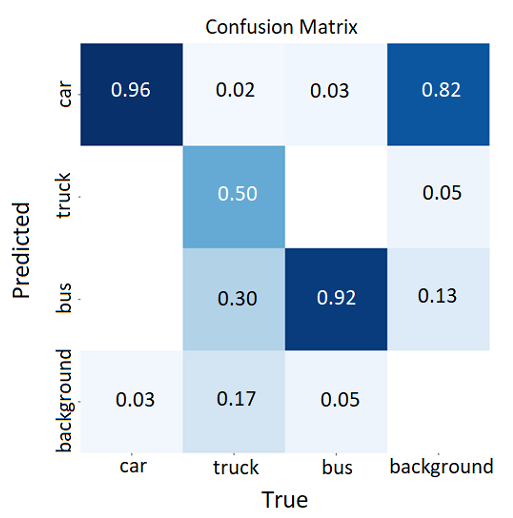}%
			\label{fig8_b}}
		\caption{Confusion matrices of (a) Single frame model, and (b) Frame pair and difference model.}
		\label{fig_sim}
	\end{figure*}
	
	\begin{figure}[!t]
		\centering
		\includegraphics[width=\columnwidth]{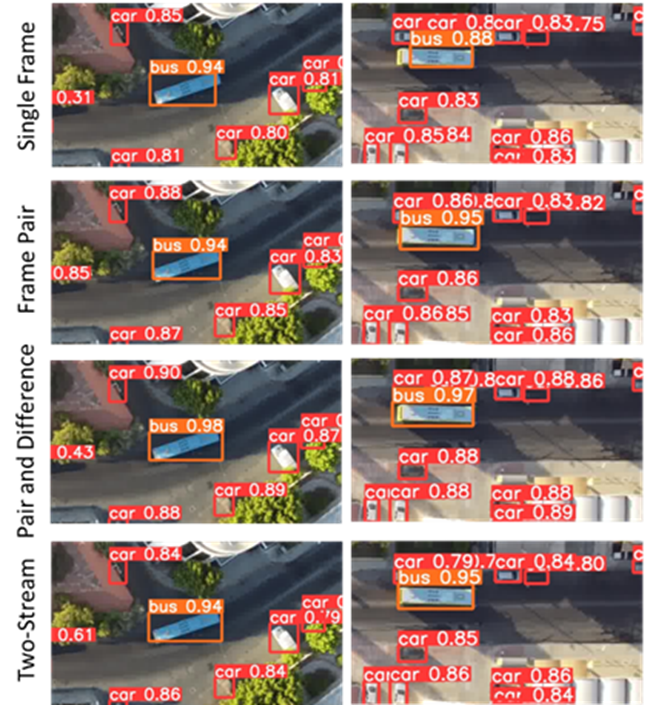}
		\caption{Inference tests regions of interest comparison for spatiotemporal models.}
		\label{fig_9}
	\end{figure}
	
	The results of attention-frame pair and difference models in Table \ref{tab:table5} show that the attention mechanisms provide a general boost in the overall mAP50, with the frame pair and difference – SE model giving a higher boost at 0.844 mAP50. However, class-specific results show that while the attention mechanisms increase the performance on the ‘bus’ class, it can also be seen that the performance of the ‘truck’ class decreases significantly in both attention models.
	
	While the attention-enhanced models indicate an overall positive effect on performance, class specific metrics indicate that this might be due to a boost in the results of one of the minority classes, however while decreasing the results of the other minority class. A possible reason for this behavior could be attributed to the fact that the model could be misclassifying the ‘truck’ and ‘bus’ classes in some cases, leading to this behaviour.
	
	A further analysis of the confusion matrices of the models in Fig. \ref{fig8_a} and Fig. \ref{fig8_b} reveals that there is indeed confusion mainly with the truck class. The single frame model misses a large portion of the dataset’s trucks, as indicated by confusion with the background for 35\% of truck instances as shown in Fig. \ref{fig8_a}. There was also 10\% and 12\% confusion with buses and cars respectively. Buses, on the other hand, were not confused by the model as trucks at all.
	
	Analysis of the confusion matrix of the frame pair and difference model in Fig. \ref{fig8_b} reveals the decreased confusion of trucks with the background and with cars, indicating the motion information significantly helps the model with identifying the presence of the trucks. However, the confusion with buses is significantly increased to 30\%, affirming the behavior of the attention-frame pair and difference models with the alternating performance between the two classes.

	\subsection{Inference Results}
	\subsubsection{Spatiotemporal Models}
	Fig. \ref{fig_9} displays the inference test results of the single frame and the three spatiotemporal models focused on regions of interest containing the minority truck and bus classes to analyze each model’s performance.
	
	\begin{figure}[!t]
		\centering
		\includegraphics[width=\columnwidth]{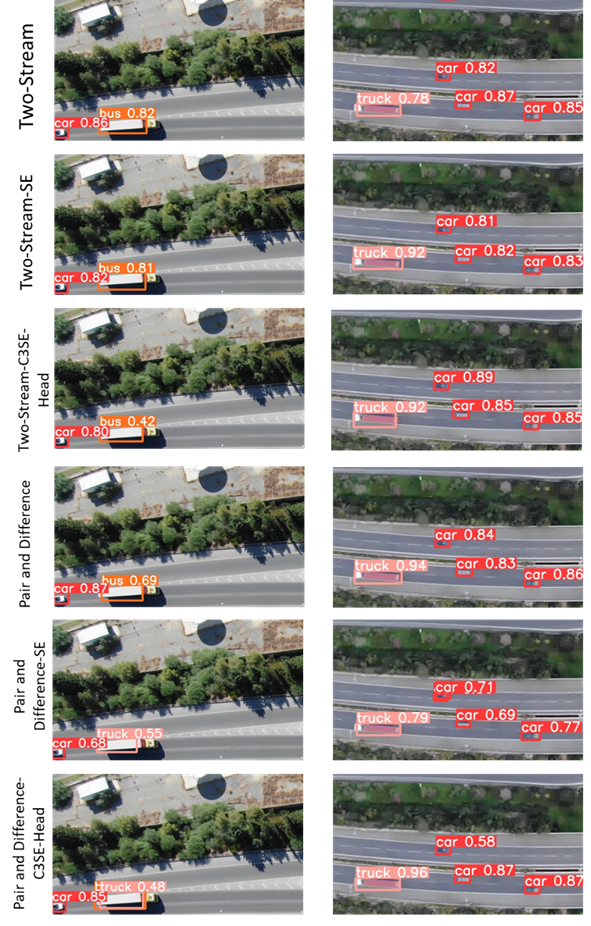}
		\caption{Inference tests regions of interest comparison for attention-spatiotemporal models.}
		\label{fig_10}
	\end{figure}
	
	\begin{figure}[!t]
		\centering
		\includegraphics[width=3.5in]{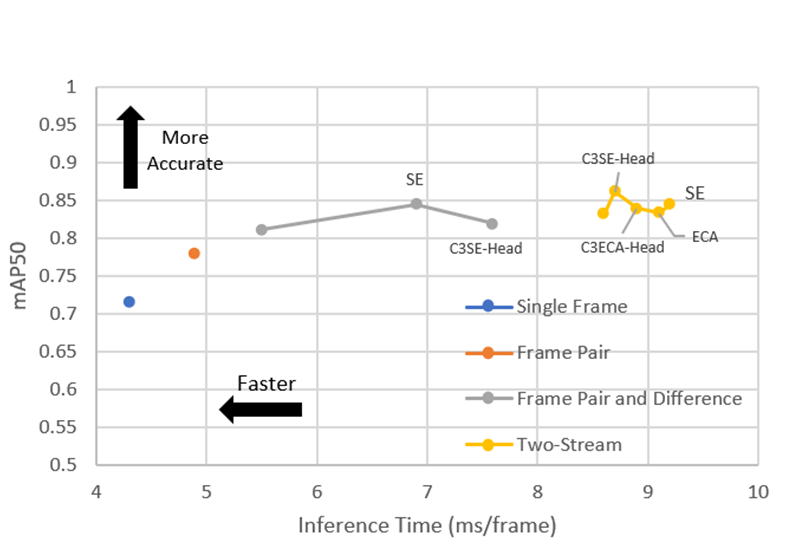}
		\caption{mAP50/Inference time for all models}
		\label{fig_11}
	\end{figure}
	
	It can be observed from Fig. \ref{fig_9} that the spatiotemporal models have a slightly higher confidence score for the 'car' class detections than the single frame model. From the first column of tests, it can be observed that the spatiotemporal models do a better job of localizing the bus than the single frame model. In the second column, the single frame model fails to correctly draw the bounding box around the bus, possibly due to the shadows cast by the nearby buildings and trees. The spatiotemporal models on the other hand show better bounding box localization on the bus with higher confidence scores of 0.95 and 0.96 even in the presence of shadowing. 
	
	Given the qualitative analysis, it can be concluded that the spatiotemporal models show better performance in localizing the minority classes, as well as better localizing and confidence scores in cases of occlusion and shadowing. The spatiotemporal models give better overall confidence scores for cars, with the frame pair and difference model showing the highest confidence scores on average. Both the single frame and the spatiotemporal models display that they can mistake semi-trucks’ trailers for buses and missing the truck detection altogether, confirming the analysis of the confusion matrices. 
	
	\subsubsection{Attention-Spatiotemporal Models}
	Fig. \ref{fig_10} displays some focused regions of interest from the inference tests on selected attention-spatiotemporal models, compared with their standard models without attention. In Fig. \ref{fig_10}, the first column reveals that both attention-two-stream models also misclassify the trailer of the semi-truck as a bus. The attention-frame pair and difference models however both indicate a positive learning response as they correctly classify the same semi-truck, although with incorrect bounding box localization, they prove that they are able to focus on features at different scales to learn to better recognize the truck. The second column of the inference tests reveals that both attention-two-stream models exhibit better localization and higher confidence scores for the truck. The frame pair and difference-C3SE-head model also exhibits the same improved behavior on the truck class, while the frame pair and difference-SE model does not. 
	
	\subsubsection{Performance/Speed Analysis}
	Inference time for all models was recorded in milliseconds per frame during the inference tests to observe the performance/speed trade-off. Fig. \ref{fig_11} shows a mAP50/Inference time graph for all models, including the single frame model, the spatiotemporal models, and their attention-infused versions. It can be observed from Fig. \ref{fig_11} that all spatiotemporal models have higher inference time than the single frame model. Nevertheless, the inference speeds were not higher by a large margin, especially in the frame pair model and frame pair and difference model where inference speeds were only 4.9 ms and 5.5 ms respectively, compared to 4.3 ms in the single frame model. The attention models of the frame pair and difference model both have considerably larger inference times than the standard model. Whereas the attention models of the two-stream model had slightly larger inference times than that of the standard two-stream model. Given the graph, it was observed that the frame pair model offers a relatively large boost in mAP50 considering the very small increase in inference time of 0.6 ms. The frame pair and difference model offers the best trade-off between inference time and performance among spatiotemporal models.
	
	\section{Discussion}
	The proposed approach holds promise for implementation across various streaming application scenarios, provided there is temporal coherency between successive frames. Despite a small computational cost, our results demonstrate that integrating temporal information can enhance accuracy compared to spatial-only models. However, the trade-off between computational efficiency and performance improvement may not always be viable, particularly considering the poorer performance on minority classes, for crucial real-world decision-making scenarios requiring near-perfect accuracy.
	
	Our work has significant potential impact, particularly in enabling drone-based traffic monitoring in areas lacking infrastructure or with specific measurements. Drones offer flexibility in adjusting positions and sampling strategies, particularly beneficial in dynamic and uncertain environments. Traffic engineering and planning authorities stand to benefit from the valuable data extracted, aiding in traffic control decisions such as vehicle counts, turning ratios, and queue lengths. Assessing the impact of false positives and negatives is vital for both micro and macro management. While minor discrepancies may be acceptable for aggregate information in traffic network control, critical applications such as traffic light control demand higher accuracy to avoid potential consequences of missed detections. Overall, this research contributes to the advancement of traffic monitoring systems by providing a way to extract real-time actionable insights for both micro and macro management strategies.
	
	\section{Conclusions and Future Work}
	This paper introduces a novel dataset for investigating spatiotemporal models for aerial vehicle detection in traffic monitoring applications. It was demonstrated that additional modified architectures and input structures that add temporal context can significantly improve detection performance. It was also observed that utilizing additional motion information in the form of frame difference in the same input stream can greatly increase overall performance with a small computational overhead over utilizing just a sequence of paired images. Furthermore, two attention mechanisms were embedded into the architecture of the spatiotemporal models at different levels. It was observed from quantitative and qualitative results that attention mechanisms indeed have the potential to enhance the learning ability of the spatiotemporal models, especially in cases of minority classes. 
	
	For future work, the spatiotemporal models ought to be experimented with a larger dataset containing more trucks and buses of all varieties to improve learning and achieve equally high accuracy in all classes. This necessitates the acquisition of more diverse data to address this issue effectively. Leveraging photo-realistic simulators offers a promising avenue to generate targeted data for underrepresented cases. Temporal augmentation techniques that can be applied to temporal data without altering spatiotemporal dependencies to improve model learning are also worth studying and investigating. Finally, this work highlights the need for further investigation into transformer models for traffic monitoring applications.
	
	\section*{Acknowledgements}
	This work is supported by the European Union Horizon 2020 Teaming, KIOS CoE, No. 739551 and from the Government of the Republic of Cyprus through the Deputy Ministry of Research, Innovation, and Digital Policy.\\
	C. Kyrkou gratefully acknowledges the support of NVIDIA Corporation with the donation of a GPU.

	\bibliographystyle{IEEEtran}  
	\bibliography{bibl}

\begin{thebibliography}{10}
\providecommand{\url}[1]{#1}
\csname url@samestyle\endcsname
\providecommand{\newblock}{\relax}
\providecommand{\bibinfo}[2]{#2}
\providecommand{\BIBentrySTDinterwordspacing}{\spaceskip=0pt\relax}
\providecommand{\BIBentryALTinterwordstretchfactor}{4}
\providecommand{\BIBentryALTinterwordspacing}{\spaceskip=\fontdimen2\font plus
\BIBentryALTinterwordstretchfactor\fontdimen3\font minus
  \fontdimen4\font\relax}
\providecommand{\BIBforeignlanguage}[2]{{%
\expandafter\ifx\csname l@#1\endcsname\relax
\typeout{** WARNING: IEEEtran.bst: No hyphenation pattern has been}%
\typeout{** loaded for the language `#1'. Using the pattern for}%
\typeout{** the default language instead.}%
\else
\language=\csname l@#1\endcsname
\fi
#2}}
\providecommand{\BIBdecl}{\relax}
\BIBdecl

\bibitem{girshick_2014_rich}
\BIBentryALTinterwordspacing
R.~Girshick, J.~Donahue, T.~Darrell, and J.~Malik, ``Rich feature hierarchies
  for accurate object detection and semantic segmentation,'' \emph{2014 IEEE
  Conference on Computer Vision and Pattern Recognition}, 06 2014. [Online].
  Available: \url{https://dl.acm.org/citation.cfm?id=2679851}
\BIBentrySTDinterwordspacing

\bibitem{girshick_2015_fast}
\BIBentryALTinterwordspacing
R.~Girshick, ``{{Fast R-CNN}},'' \emph{2015 IEEE International Conference on
  Computer Vision (ICCV)}, pp. 1440--1448, 12 2015. [Online]. Available:
  \url{https://ieeexplore.ieee.org/document/7410526}
\BIBentrySTDinterwordspacing

\bibitem{ren_2017_faster}
\BIBentryALTinterwordspacing
S.~Ren, K.~He, R.~Girshick, and J.~Sun, ``{{Faster R-CNN: Towards Real-Time
  Object Detection with Region Proposal Networks}},'' \emph{IEEE Transactions
  on Pattern Analysis and Machine Intelligence}, vol.~39, pp. 1137--1149, 06
  2017. [Online]. Available: \url{https://ieeexplore.ieee.org/document/7485869}
\BIBentrySTDinterwordspacing

\bibitem{redmon_2016_you}
\BIBentryALTinterwordspacing
J.~Redmon, S.~Divvala, R.~Girshick, and A.~Farhadi, ``You only look once:
  Unified, real-time object detection,'' \emph{2016 IEEE Conference on Computer
  Vision and Pattern Recognition (CVPR)}, 2016. [Online]. Available:
  \url{https://arxiv.org/pdf/1506.02640.pdf}
\BIBentrySTDinterwordspacing

\bibitem{redmon_2017_yolo9000}
\BIBentryALTinterwordspacing
J.~Redmon and A.~Farhadi, ``{{YOLO9000: Better, Faster, Stronger}},''
  \emph{2017 IEEE Conference on Computer Vision and Pattern Recognition
  (CVPR)}, 07 2017. [Online]. Available:
  \url{https://ieeexplore.ieee.org/document/8100173}
\BIBentrySTDinterwordspacing

\bibitem{redmon_2018_yolov3}
\BIBentryALTinterwordspacing
------, ``{{YOLOv3: An Incremental Improvement}},'' \emph{arXiv.org}, 2018.
  [Online]. Available: \url{https://arxiv.org/abs/1804.02767}
\BIBentrySTDinterwordspacing

\bibitem{bochkovskiy_2020_yolov4}
\BIBentryALTinterwordspacing
A.~Bochkovskiy, C.-Y. Wang, and H.-Y.~M. Liao, ``{{YOLOv4: Optimal Speed and
  Accuracy of Object Detection}},'' arXiv.org, 04 2020. [Online]. Available:
  \url{https://arxiv.org/abs/2004.10934}
\BIBentrySTDinterwordspacing

\bibitem{jocher_2022_ultralyticsyolov5}
\BIBentryALTinterwordspacing
G.~Jocher, ``{YOLOv5 by Ultralytics},'' May 2020. [Online]. Available:
  \url{https://github.com/ultralytics/yolov5}
\BIBentrySTDinterwordspacing

\bibitem{li2022yolov6}
C.~Li, L.~Li, H.~Jiang, K.~Weng, Y.~Geng, L.~Li, Z.~Ke, Q.~Li, M.~Cheng, W.~Nie
  \emph{et~al.}, ``{{YOLOv6: A single-stage object detection framework for
  industrial applications}},'' \emph{arXiv preprint arXiv:2209.02976}, 2022.

\bibitem{wang2023yolov7}
C.-Y. Wang, A.~Bochkovskiy, and H.-Y.~M. Liao, ``{{YOLOv7: Trainable
  bag-of-freebies sets new state-of-the-art for real-time object detectors}},''
  in \emph{Proceedings of the IEEE/CVF conference on computer vision and
  pattern recognition}, 2023, pp. 7464--7475.

\bibitem{yoloreview}
J.~Terven, D.-M. Córdova-Esparza, and J.-A. Romero-González, ``{{A
  Comprehensive Review of YOLO Architectures in Computer Vision: From YOLOv1 to
  YOLOv8 and YOLO-NAS}},'' \emph{Machine Learning and Knowledge Extraction},
  vol.~5, no.~4, pp. 1680--1716, 2023.

\bibitem{carion2020end}
N.~Carion, F.~Massa, G.~Synnaeve, N.~Usunier, A.~Kirillov, and S.~Zagoruyko,
  ``End-to-end object detection with transformers,'' in \emph{European
  conference on computer vision}.\hskip 1em plus 0.5em minus 0.4em\relax
  Springer, 2020, pp. 213--229.

\bibitem{lv2023detrs}
W.~Lv, S.~Xu, Y.~Zhao, G.~Wang, J.~Wei, C.~Cui, Y.~Du, Q.~Dang, and Y.~Liu,
  ``{{DETRs Beat YOLOs on Real-time Object Detection}},'' 2023.

\bibitem{fang2021you}
Y.~Fang, B.~Liao, X.~Wang, J.~Fang, J.~Qi, R.~Wu, J.~Niu, and W.~Liu, ``You
  only look at one sequence: Rethinking transformer in vision through object
  detection,'' \emph{Advances in Neural Information Processing Systems},
  vol.~34, pp. 26\,183--26\,197, 2021.

\bibitem{Pflugfelder2017SiameseVisualTracking}
\BIBentryALTinterwordspacing
R.~P. Pflugfelder, ``Siamese learning visual tracking: {A} survey,''
  \emph{CoRR}, vol. abs/1707.00569, 2017. [Online]. Available:
  \url{http://arxiv.org/abs/1707.00569}
\BIBentrySTDinterwordspacing

\bibitem{han_2016_seqnms}
\BIBentryALTinterwordspacing
W.~Han, P.~Khorrami, T.~Le~Paine, P.~Ramachandran, M.~Babaeizadeh, H.~Shi,
  J.~Li, S.~Yan, and T.~Huang, ``{{Seq-NMS for Video Object Detection}},''
  2016. [Online]. Available: \url{https://arxiv.org/pdf/1602.08465.pdf}
\BIBentrySTDinterwordspacing

\bibitem{kang_2017_object}
K.~Kang, H.~Li, T.~Xiao, W.~Ouyang, J.~Yan, X.~Liu, and X.~Wang, ``{{Object
  Detection in Videos with Tubelet Proposal Networks}},'' \emph{2017 IEEE
  Conference on Computer Vision and Pattern Recognition (CVPR)}, 07 2017.

\bibitem{kang_2018_tcnn}
\BIBentryALTinterwordspacing
K.~Kang, H.~Li, J.~Yan, X.~Zeng, B.~Yang, T.~Xiao, C.~Zhang, Z.~Wang, R.~Wang,
  X.~Wang, and W.~Ouyang, ``{{T-CNN: Tubelets with Convolutional Neural
  Networks for Object Detection from Videos}},'' \emph{IEEE Transactions on
  Circuits and Systems for Video Technology}, vol.~28, p. 2896–2907, 10 2018.
  [Online]. Available: \url{https://arxiv.org/abs/1604.02532}
\BIBentrySTDinterwordspacing

\bibitem{zhu_2017_flowguided}
X.~Zhu, Y.~Wang, J.~Dai, L.~Yuan, and Y.~Wei, ``Flow-guided feature aggregation
  for video object detection,'' \emph{2017 IEEE International Conference on
  Computer Vision (ICCV)}, 10 2017.

\bibitem{zhu_2017_deep}
\BIBentryALTinterwordspacing
X.~Zhu, Y.~Xiong, J.~Dai, L.~Yuan, and Y.~Wei, ``Deep feature flow for video
  recognition,'' \emph{2017 IEEE Conference on Computer Vision and Pattern
  Recognition (CVPR)}, 2017. [Online]. Available:
  \url{https://www.semanticscholar.org/paper/Deep-Feature-Flow-for-Video-Recognition-Zhu-Xiong/5c544788faa5b6031db5020bbdaeb25e68c24e19}
\BIBentrySTDinterwordspacing

\bibitem{shi_2015_convolutional}
\BIBentryALTinterwordspacing
X.~Shi, Z.~Chen, H.~Wang, D.-Y. Yeung, W.-K. Wong, W.-C. Woo, and
  H.~Kong~Observatory, ``{{Convolutional LSTM Network: A Machine Learning
  Approach for Precipitation Nowcasting}},'' 2015. [Online]. Available:
  \url{https://proceedings.neurips.cc/paper/2015/file/07563a3fe3bbe7e3ba84431ad9d055af-Paper.pdf}
\BIBentrySTDinterwordspacing

\bibitem{lu_2017_online}
Y.~Lu, C.~Lu, and C.-K. Tang, ``Online video object detection using association
  lstm,'' \emph{International Conference on Computer Vision}, 10 2017.

\bibitem{ICDSC2018plastirasSTP}
\BIBentryALTinterwordspacing
G.~Plastiras, C.~Kyrkou, and T.~Theocharides, ``{Efficient ConvNet-Based Object
  Detection for Unmanned Aerial Vehicles by Selective Tile Processing},'' in
  \emph{Proceedings of the 12th International Conference on Distributed Smart
  Cameras}, ser. ICDSC '18.\hskip 1em plus 0.5em minus 0.4em\relax New York,
  NY, USA: Association for Computing Machinery, 2018. [Online]. Available:
  \url{https://doi.org/10.1145/3243394.3243692}
\BIBentrySTDinterwordspacing

\bibitem{Makrigiorgis2022}
R.~Makrigiorgis, N.~Hadjittoouli, C.~Kyrkou, and T.~Theocharides,
  ``{{AirCamRTM: Enhancing Vehicle Detection for Efficient Aerial Camera-based
  Road Traffic Monitoring}},'' in \emph{2022 IEEE/CVF Winter Conference on
  Applications of Computer Vision (WACV)}, 2022, pp. 3431--3440.

\bibitem{Kyrkou2018}
C.~Kyrkou, S.~Timotheou, P.~Kolios, T.~Theocharides, and C.~G. Panayiotou,
  ``{{Optimized vision-directed deployment of UAVs for rapid traffic
  monitoring}},'' in \emph{2018 IEEE International Conference on Consumer
  Electronics (ICCE)}, 2018, pp. 1--6.

\bibitem{KyrkouPotentials2019}
C.~Kyrkou, S.~Timotheou, P.~Kolios, T.~Theocharides, and C.~Panayiotou,
  ``{{Drones: Augmenting Our Quality of Life}},'' \emph{IEEE Potentials},
  vol.~38, no.~1, pp. 30--36, 2019.

\bibitem{liu_2016_ssd}
\BIBentryALTinterwordspacing
W.~Liu, D.~Anguelov, D.~Erhan, C.~Szegedy, S.~Reed, C.-Y. Fu, and A.~C. Berg,
  ``{{SSD: Single Shot MultiBox Detector}},'' \emph{Computer Vision – ECCV
  2016}, pp. 21--37, 2016. [Online]. Available:
  \url{https://arxiv.org/abs/1512.02325}
\BIBentrySTDinterwordspacing

\bibitem{ramachandran2019stand}
P.~Ramachandran, N.~Parmar, A.~Vaswani, I.~Bello, A.~Levskaya, and J.~Shlens,
  ``{{Stand-alone self-attention in vision models}},'' \emph{Advances in neural
  information processing systems}, vol.~32, 2019.

\bibitem{bello2019attention}
I.~Bello, B.~Zoph, A.~Vaswani, J.~Shlens, and Q.~V. Le, ``Attention augmented
  convolutional networks,'' in \emph{Proceedings of the IEEE/CVF international
  conference on computer vision}, 2019, pp. 3286--3295.

\bibitem{khan2023survey}
A.~Khan, Z.~Rauf, A.~Sohail, A.~R. Khan, H.~Asif, A.~Asif, and U.~Farooq, ``{A
  survey of the vision transformers and their CNN-transformer based
  variants},'' \emph{Artificial Intelligence Review}, vol.~56, no. Suppl 3, pp.
  2917--2970, 2023.

\bibitem{jiao_2021_new}
L.~Jiao, R.~Zhang, F.~Liu, S.~Yang, B.~Hou, L.~Li, and X.~Tang, ``{New
  Generation Deep Learning for Video Object Detection: A Survey},'' \emph{IEEE
  Transactions on Neural Networks and Learning Systems}, pp. 1--21, 2021.

\bibitem{hochreiter_1997_long}
S.~Hochreiter and J.~Schmidhuber, ``{{Long Short-Term Memory}},'' \emph{Neural
  Computation}, vol.~9, pp. 1735--1780, 11 1997.

\bibitem{liu2018mobile}
M.~Liu and M.~Zhu, ``{{Mobile video object detection with temporally-aware
  feature maps}},'' in \emph{Proceedings of the IEEE conference on computer
  vision and pattern recognition}, 2018, pp. 5686--5695.

\bibitem{zhang2019modeling}
C.~Zhang and J.~Kim, ``{{Modeling long-and short-term temporal context for
  video object detection}},'' in \emph{2019 IEEE international conference on
  image processing (ICIP)}.\hskip 1em plus 0.5em minus 0.4em\relax IEEE, 2019,
  pp. 71--75.

\bibitem{deng2019object}
H.~Deng, Y.~Hua, T.~Song, Z.~Zhang, Z.~Xue, R.~Ma, N.~Robertson, and H.~Guan,
  ``{{Object guided external memory network for video object detection}},'' in
  \emph{Proceedings of the IEEE/CVF International Conference on Computer
  Vision}, 2019, pp. 6678--6687.

\bibitem{beery2020context}
S.~Beery, G.~Wu, V.~Rathod, R.~Votel, and J.~Huang, ``{{Context R-CNN: Long
  term temporal context for per-camera object detection}},'' in
  \emph{Proceedings of the IEEE/CVF conference on computer vision and pattern
  recognition}, 2020, pp. 13\,075--13\,085.

\bibitem{ji_2013_3d}
\BIBentryALTinterwordspacing
S.~Ji, W.~Xu, M.~Yang, and K.~Yu, ``{{3D Convolutional Neural Networks for
  Human Action Recognition}},'' \emph{IEEE Transactions on Pattern Analysis and
  Machine Intelligence}, vol.~35, pp. 221--231, 01 2013. [Online]. Available:
  \url{http://users.eecs.northwestern.edu/~mya671/mypapers/ICML10_Ji_Xu_Yang_Yu.pdf}
\BIBentrySTDinterwordspacing

\bibitem{tran_2015_learning}
\BIBentryALTinterwordspacing
D.~Tran, L.~Bourdev, R.~Fergus, L.~Torresani, and M.~Paluri, ``{{Learning
  Spatiotemporal Features with 3D Convolutional Networks}},'' \emph{2015 IEEE
  International Conference on Computer Vision (ICCV)}, 12 2015. [Online].
  Available:
  \url{https://nyuscholars.nyu.edu/en/publications/learning-spatiotemporal-features-with-3d-convolutional-networks}
\BIBentrySTDinterwordspacing

\bibitem{lin_2020_tsm}
J.~Lin, C.~Gan, K.~Wang, and S.~Han, ``Tsm: Temporal shift module for efficient
  and scalable video understanding on edge devices,'' \emph{IEEE Transactions
  on Pattern Analysis and Machine Intelligence}, pp. 1--1, 2020.

\bibitem{PASSOS2022101754}
W.~L. Passos, G.~M. Araujo, A.~A. {de Lima}, S.~L. Netto, and E.~A. {da Silva},
  ``{{Automatic detection of Aedes aegypti breeding grounds based on deep
  networks with spatio-temporal consistency}},'' \emph{Computers, Environment
  and Urban Systems}, vol.~93, p. 101754, 2022.

\bibitem{lalonde_2018_clusternet}
\BIBentryALTinterwordspacing
R.~LaLonde, D.~Zhang, and M.~Shah, ``{{ClusterNet: Detecting Small Objects in
  Large Scenes by Exploiting Spatio-Temporal Information}},'' IEEE Xplore, p.
  4003–4012, 06 2018. [Online]. Available:
  \url{https://ieeexplore.ieee.org/stamp/stamp.jsp?tp=&arnumber=8578519}
\BIBentrySTDinterwordspacing

\bibitem{corsel_2023_exploiting}
\BIBentryALTinterwordspacing
C.~W. Corsel, M.~van Lier, L.~Kampmeijer, N.~Boehrer, and E.~M. Bakker,
  ``{{Exploiting Temporal Context for Tiny Object Detection}},'' IEEE Xplore,
  p. 1–11, 01 2023. [Online]. Available:
  \url{https://ieeexplore.ieee.org/stamp/stamp.jsp?tp=&arnumber=10031105}
\BIBentrySTDinterwordspacing

\bibitem{niu_2021_a}
Z.~Niu, G.~Zhong, and H.~Yu, ``A review on the attention mechanism of deep
  learning,'' \emph{Neurocomputing}, vol. 452, pp. 48--62, 09 2021.

\bibitem{guo_2022_attention}
M.-H. Guo, T.-X. Xu, J.-J. Liu, Z.-N. Liu, P.-T. Jiang, T.-J. Mu, S.-H. Zhang,
  R.~R. Martin, M.-M. Cheng, and S.-M. Hu, ``{{Attention mechanisms in computer
  vision: A survey}},'' \emph{Computational Visual Media}, vol.~8, 03 2022.

\bibitem{mnih_2014_recurrent}
\BIBentryALTinterwordspacing
V.~Mnih, N.~Heess, A.~Graves, K.~Kavukcuoglu, and G.~Deepmind, ``Recurrent
  models of visual attention,'' 2014. [Online]. Available:
  \url{https://proceedings.neurips.cc/paper_files/paper/2014/file/09c6c3783b4a70054da74f2538ed47c6-Paper.pdf}
\BIBentrySTDinterwordspacing

\bibitem{jaderberg_2015_spatial}
M.~Jaderberg, K.~Simonyan, A.~Zisserman \emph{et~al.}, ``Spatial transformer
  networks,'' \emph{Advances in neural information processing systems},
  vol.~28, 2015.

\bibitem{hu_2019_squeezeandexcitation}
J.~Hu, L.~Shen, S.~Albanie, G.~Sun, and E.~Wu, ``{{Squeeze-and-Excitation
  Networks}},'' \emph{IEEE Transactions on Pattern Analysis and Machine
  Intelligence}, pp. 1--1, 2019.

\bibitem{wang_2020_ecanet}
Q.~Wang, B.~Wu, P.~Zhu, P.~Li, W.~Zuo, and Q.~Hu, ``{{ECA-Net: Efficient
  Channel Attention for Deep Convolutional Neural Networks}},'' \emph{2020
  IEEE/CVF Conference on Computer Vision and Pattern Recognition (CVPR)}, 06
  2020.

\bibitem{woo_2018_cbam}
S.~Woo, J.~Park, J.-Y. Lee, and I.~S. Kweon, ``{{CBAM: Convolutional Block
  Attention Module}},'' \emph{Computer Vision – ECCV 2018}, pp. 3--19, 2018.

\bibitem{vaswani_2017_attention}
\BIBentryALTinterwordspacing
A.~Vaswani, N.~Shazeer, N.~Parmar, J.~Uszkoreit, L.~Jones, A.~N. Gomez,
  L.~Kaiser, and I.~Polosukhin, ``Attention is all you need,'' \emph{Neural
  Information Processing Systems}, vol.~30, 2017. [Online]. Available:
  \url{https://proceedings.neurips.cc/paper_files/paper/2017/hash/3f5ee243547dee91fbd053c1c4a845aa-Abstract.html}
\BIBentrySTDinterwordspacing

\bibitem{wang_2018_nonlocal}
X.~Wang, R.~Girshick, A.~Gupta, and K.~He, ``Non-local neural networks,''
  \emph{2018 IEEE/CVF Conference on Computer Vision and Pattern Recognition},
  06 2018.

\bibitem{dosovitskiy_2020_an}
\BIBentryALTinterwordspacing
A.~Dosovitskiy, L.~Beyer, A.~Kolesnikov, D.~Weissenborn, X.~Zhai,
  T.~Unterthiner, M.~Dehghani, M.~Minderer, G.~Heigold, S.~Gelly, J.~Uszkoreit,
  and N.~Houlsby, ``An image is worth 16x16 words: Transformers for image
  recognition at scale,'' \emph{arXiv:2010.11929 [cs]}, 10 2020. [Online].
  Available: \url{https://arxiv.org/abs/2010.11929}
\BIBentrySTDinterwordspacing

\bibitem{wang2021pyramid}
W.~Wang, E.~Xie, X.~Li, D.-P. Fan, K.~Song, D.~Liang, T.~Lu, P.~Luo, and
  L.~Shao, ``Pyramid vision transformer: A versatile backbone for dense
  prediction without convolutions,'' in \emph{Proceedings of the IEEE/CVF
  international conference on computer vision}, 2021, pp. 568--578.

\bibitem{liu2021swin}
Z.~Liu, Y.~Lin, Y.~Cao, H.~Hu, Y.~Wei, Z.~Zhang, S.~Lin, and B.~Guo, ``Swin
  transformer: Hierarchical vision transformer using shifted windows,'' in
  \emph{Proceedings of the IEEE/CVF international conference on computer
  vision}, 2021, pp. 10\,012--10\,022.

\bibitem{bertasius2021space}
G.~Bertasius, H.~Wang, and L.~Torresani, ``Is space-time attention all you need
  for video understanding?'' in \emph{ICML}, vol.~2, no.~3, 2021, p.~4.

\bibitem{arnab_2021_vivit}
\BIBentryALTinterwordspacing
A.~Arnab, M.~Dehghani, G.~Heigold, C.~Sun, M.~Lučić, and C.~Schmid, ``Vivit:
  A video vision transformer,'' IEEE Xplore, p. 6816–6826, 10 2021. [Online].
  Available: \url{https://ieeexplore.ieee.org/abstract/document/9710415}
\BIBentrySTDinterwordspacing

\bibitem{cao2021visdrone}
Y.~Cao, Z.~He, L.~Wang, W.~Wang, Y.~Yuan, D.~Zhang, J.~Zhang, P.~Zhu,
  L.~Van~Gool, J.~Han \emph{et~al.}, ``{{VisDrone-DET2021: The vision meets
  drone object detection challenge results}},'' in \emph{Proceedings of the
  IEEE/CVF International conference on computer vision}, 2021, pp. 2847--2854.

\bibitem{DroneVehicle}
Y.~Sun, B.~Cao, P.~Zhu, and Q.~Hu, ``{{Drone-Based RGB-Infrared Cross-Modality
  Vehicle Detection Via Uncertainty-Aware Learning}},'' \emph{IEEE Transactions
  on Circuits and Systems for Video Technology}, vol.~32, no.~10, pp.
  6700--6713, 2022.

\bibitem{Makrigiorgis2023}
R.~Makrigiorgis, C.~Kyrkou, and P.~Kolios, ``How high can you detect? improved
  accuracy and efficiency at varying altitudes for aerial vehicle detection,''
  in \emph{2023 International Conference on Unmanned Aircraft Systems (ICUAS)},
  2023, pp. 167--174.

\bibitem{wang_2020_cspnet}
C.-Y. Wang, H.-Y. Mark~Liao, Y.-H. Wu, P.-Y. Chen, J.-W. Hsieh, and I.-H. Yeh,
  ``{{CSPNet: A New Backbone that can Enhance Learning Capability of CNN}},''
  \emph{2020 IEEE/CVF Conference on Computer Vision and Pattern Recognition
  Workshops (CVPRW)}, 06 2020.

\bibitem{padilla_2021_a}
R.~Padilla, W.~L. Passos, T.~L.~B. Dias, S.~L. Netto, and E.~A.~B. da~Silva,
  ``A comparative analysis of object detection metrics with a companion
  open-source toolkit,'' \emph{Electronics}, vol.~10, p. 279, 01 2021.

\bibitem{padilla_2020_a}
R.~Padilla, S.~L. Netto, and E.~A.~B. da~Silva, ``A survey on performance
  metrics for object-detection algorithms,'' \emph{2020 International
  Conference on Systems, Signals and Image Processing (IWSSIP)}, 07 2020.

\end{thebibliography}
\end{document}